\documentclass[letterpaper, 10 pt, conference]{ieeeconf}  % Comment this line out if you need a4paper

\IEEEoverridecommandlockouts                              % This command is only needed if 
\usepackage{amsthm}
\usepackage{graphicx}
\usepackage{times}
\usepackage{soul}
\usepackage{url}
\usepackage[utf8]{inputenc}
\usepackage[small]{caption}
\usepackage{booktabs}
\usepackage{algorithm}
\usepackage[switch]{lineno}
\usepackage{amsmath,amssymb,amsfonts}
\usepackage{float}
\usepackage[table,xcdraw]{xcolor}
\usepackage{textcomp}
\usepackage{hyperref}
\usepackage{algpseudocode}
\usepackage{multirow, multicol}
\usepackage{cite}
\usepackage{tabularx}
\title{\LARGE \bf
BVIP Guiding System with Adaptability to Individual Differences}
\author{Yibo Zhou$^{\dag}$\and
Dongfei Cui$^{\dag}$\and
Xiangming Dong$^{\dag}$\and
Zongkai Wu$^{*}$\and
Zhenyu Wei\and
Donglin Wang*
\thanks{$\dag$ Contributed equally    $*$ Co-corresponding author}
\thanks{This work is supported by the Westlake University and Hangzhou Science and Technology Bureau.}
\thanks{Dongfei Cui, Zongkai Wu, Zhenyu Wei and Donglin Wang are with are with Machine Intelligence Lab (MiLAB), School of Engineering, Westlake University.}
\thanks{Yibo Zhou is with National Elite Institute of Engineering, Chongqing University, Chongqing, China}
\thanks{Xiangming Dong is with School of Mechanical and Electrical Engineering, Wenzhou University, Wenzhou, Zhejiang Province, China}
}
\begin{document}
\maketitle
\thispagestyle{empty}
\pagestyle{empty}
%%%%%%%%%%%%%%%%%%%%%%%%%%%%%%%%%%%%%%%%%%%%%%%%%%%%%%%%%%%%%%%%%%%%%%%%%%%%%%%%
\begin{abstract}

Guiding robots can not only detect close-range obstacles like other guiding tools, but also extend its range to perceive the environment when making decisions.
However, most existing works over-simplified the interaction between human agents and robots, ignoring the differences between individuals, resulting in poor experiences for different users.
To solve the problem, we propose a data-driven guiding system to cope with the effect brighten by individual differences.
In our guiding system, we design a Human Motion Predictor (HMP) and a Robot Dynamics Model (RDM) based on deep neural network, the time convolutional network (TCN) is verified to have the best performance, to predict differences in interaction between different human agents and robots.
To train our models, we collected datasets that records the interactions from different human agents.
Moreover, given the predictive information of the specific user, we propose a waypoints selector that allows the robot to naturally adapt to the user's state changes, which are mainly reflected in the walking speed.
We compare the performance of our models with previous works and achieve significant performance improvements.
On this basis, our guiding system demonstrated good adaptability to different human agents.
Our guiding system is deployed on a real quadruped robot to verify the practicability.

\end{abstract}
%%%%%%%%%%%%%%%%%%%%%%%%%%%%%%%%%%%%%%%%%%%%%%%%%%%%%%%%%%%%%%%%%%%%%%%%%%%%%%%%
\section{INTRODUCTION}
As for blind and visually impaired people (BVIP), traveling is necessary but difficult. 
There are many studies dedicated to solving the task of guiding BVIP. The solutions include intelligent white cane\cite{li2018vision}, intelligent wearables\cite{filipe2012blind} and etc. 
Among these solutions, guiding robots can not only detect close-range obstacles like other guiding tools, but also extend its range to perceive the environment when making decisions, which contribute to its outstanding performance. 
Besides, with the excellent maneuverability and terrain adaptability\cite{gilroy2021autonomous, mastalli2020motion}, quadruped robots have become an excellent chassis for guiding robots.

\begin{figure}[t]
    \centering
    \includegraphics[width=0.48\textwidth,height=0.267\textwidth]{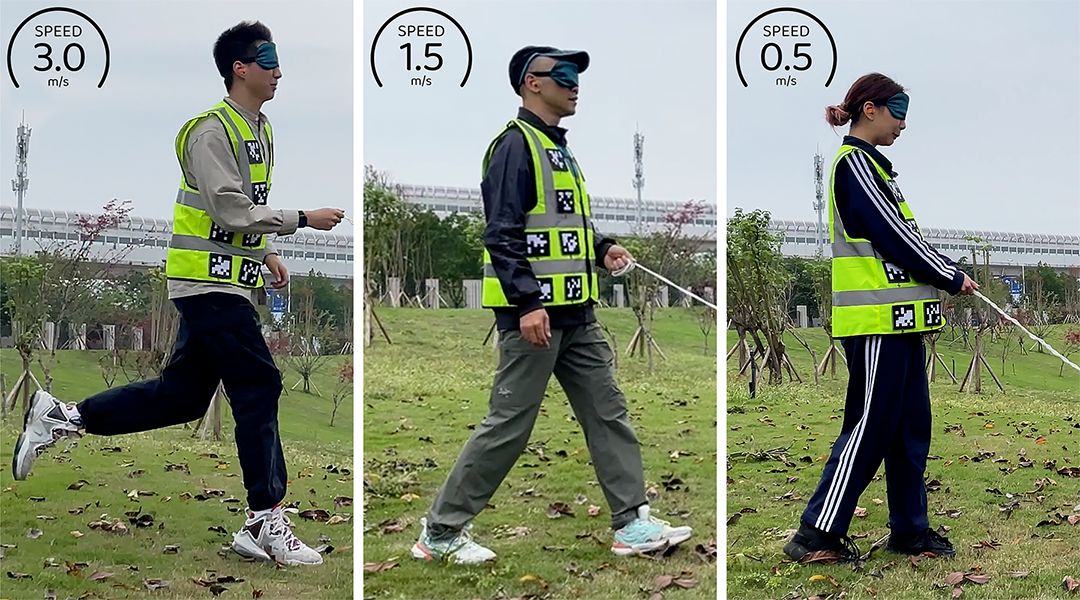}
    \caption{Different human agents are guided by our quadruped BVIP guiding robot. Each of them has their own comfortable walking speed, and our guiding system can flexibly adapt their differences.}
    \label{IndividualDifferences}
\end{figure}

However, individual differences among human agents can lead to different guiding interactions. 
For instance, a tired person may walk slower than an energetic person when being guided with same tension(see Fig. \ref{IndividualDifferences}). 
Previous researches lacked the consideration of the individual differences, resulting in stiff guiding performance. 
To eliminate individual differences, human agents were asked to walk slowly at the same speed. 
More concretely, the lack of considering individual differences leads to 2 problems: 1) the motions of both the robot and the human agent after being towed are predicted inaccurately; 2) the robot's velocity cannot adapt to different human agents' velocities.

To address the problems, we propose a data-driven BVIP guiding system to cope with the effect brighten by individual differences.
In our guiding system, we first use the position of the robot and camera positioning technology to locate the position of the human, and then plan the human route.

%%%%%%%%%%%%%%%%%%%%%%%%%%%%%%%%%%%%%%%%%%%%%%%%%%%%%%%%%%%%%%%%%%%%%%%%%%%%
\begin{figure*}[htp]
    \includegraphics[width=\textwidth]{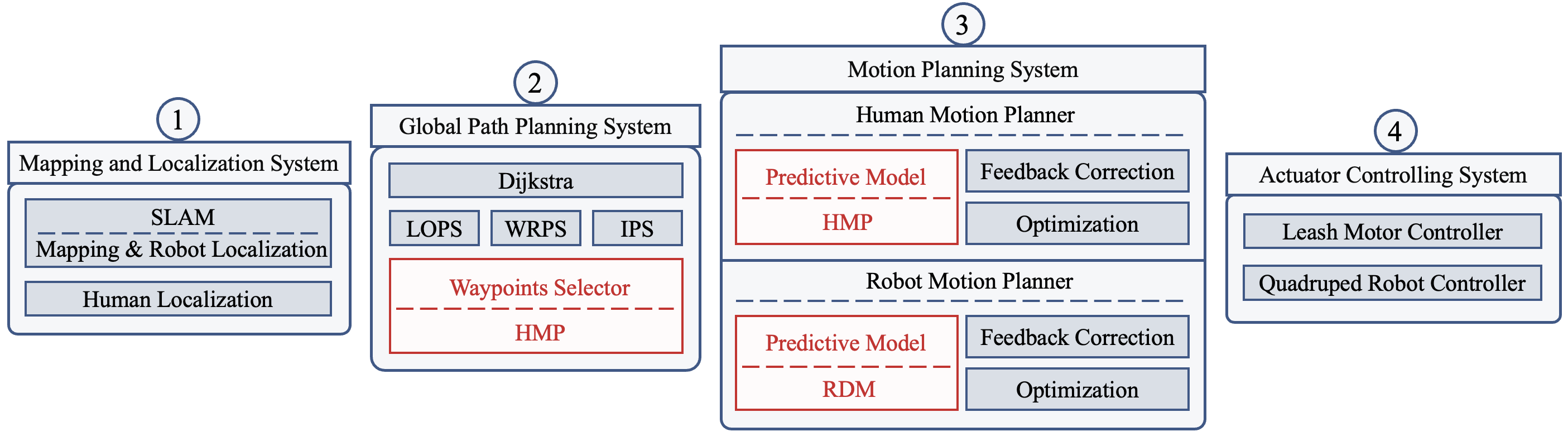}
    \caption{Overview of our BVIP guiding system.}
    \label{SystemFramework}
\end{figure*}
%%%%%%%%%%%%%%%%%%%%%%%%%%%%%%%%%%%%%%%%%%%%%%%%%%%%%%%%%%%%%%%%%%%%%%%%%%%%

Precisely predicting different human agents' motions is the basis of motion planning in BVIP guiding task.
Our experiment proved that human's velocity is related to the received tensions in past periods. 
Thus, we propose a Human Motion Predictor (HMP) that based on well-known neural architectures, which can precisely predict human agents' motions according to their received tensions.
To train the model, we collected a dataset which records the motion data of different humans after being towed.

Next, to adapt different human agents' comfortable speeds, we propose a Waypoints Selector. The selector chooses waypoints on the planned global path as the target positions for human agents. The faster the human agent walks, the greater the spacing between the selected waypoints.

Model Predictive Control (MPC) is then used to plan the leash tensions and robot's velocity which are required to make human agents follow the selected waypoints. Besides predicting the motions of human agents, we also establish a Robot Dynamics Model (RDM) to describe robot's motion responds after being towed. A dateset that record robot's responds after receiving different tensions from each directions is collected to train the model.
Our neural based models are necessary to the MPC-based motion planner.

In experiments, we compared our models with previous researches. It can be concluded that our models achieved significant performance improvements. On this basis, we deployed our BVIP guiding system on a quadruped robot. The synchronization between the robot and different human agents is tested. It is demonstrated that our BVIP guiding system has good adaptability to individual differences.
our contributions are summarized as follows,
\begin{itemize}
    \item Considering the individual differences among human agents can lead to different guiding interactions. we propose the data-driven guiding system for BVIP. In our system, In our guiding system, we design a AI-based HMP and RDM to predict differences in interaction between different human agents and robots.
    \item  We propose a waypoint selector that works with motion predictors and MPC to guide different people at a comfortable pace.
    \item On the experimental side, we first validate the performance improvements between our predictor and baselines, while analyzing the differences between three well-known architectures in our task, resulting in the factual results of TCN optimality. 
    Secondly, we also applied our system to real robots to verify its practicability.
\end{itemize}

% We are looking forward to inspire future works in a hope that blind and visually impaired groups can travel easily in the future.
%%%%%%%%%%%%%%%%%%%%%%%%%%%%%%%%%%%%%%%%%%%%%%%%%%%%%%%%%%%%%%%%%%%%%%%%%%%%%%%%
\section{RELATED WORKS}
\subsection{Human Motion Prediction}
The mainstream human motion predicting algorithms now is coupled approaches. They focused on modeling multi-agent interactions, which solves many problems shown by decoupled approaches, such as reciprocal dance problem\cite{feurtey2000simulating}. Nanavati et al.\cite{nanavati2019follow} proposed a coupled human motion predicting model for BVIP guiding task, which was based on Markov Model. However, they ignored the impact of the tension received by human agents. They also left the individual differences amoung human agents out of consideration. Although the model of Chen et al.\cite{chen2022quadruped} considered the problems above, they over-simplified the model into linear relation, which had no time-ordered. Further more, both two models were hand-made, which leads to inaccurate performance.

Herath et al.\cite{herath2020ronin} proposed a model which was based on well-known neural architectures, which predict human's position from IMU data. Their work is inspiring to our human motion predicting task.
%%%%%%%%%%%%%%%%%%%%%%%%%%%%%%%%%%%%%%%%%%%%%%%%%%%%%%%%%%%%%%%%%%%%%%%%%%%%%%%%
\subsection{Robot Dynamics Model Establishment}
Modeling specific platforms accurately is usually complicated when using model-based controlling methods. Chen et al.\cite{chen2022quadruped} over-simplified the effect of tension on the velocity of quadruped robots into a velocity discount coefficient matrix, assuming that only the force at the current moment would cause a loss in the robot's velocity. However, the model is highly inaccurate.

We are inspired by previous works in other domains. Gillespie et al.\cite{gillespie2018learning} used deep neural networks to learn the nonlinear dynamics models of a soft robot, which is difficult to manually model as quadruped robots. Spielberg et al.\cite{spielberg2021neural} applied neural networks to modeling unknown friction in automated driving task. 
%%%%%%%%%%%%%%%%%%%%%%%%%%%%%%%%%%%%%%%%%%%%%%%%%%%%%%%%%%%%%%%%%%%%%%%%%%%%%%%%
\subsection{Global Path Correction}
Traditional grid-based global path planning algorithms generate paths consisting of adjacent grid points with equal distances between them, such as A* algorithm and Dijkstra algorithm. Song et al.[9] proposed an algorithm called smoothed A*, which eliminates the redundant points arise in the traditional A* algorithm. They also used interpolation to convert the discrete path into an analytical path. Our algorithm is further improved upon this foundation.
%%%%%%%%%%%%%%%%%%%%%%%%%%%%%%%%%%%%%%%%%%%%%%%%%%%%%%%%%%%%%%%%%%%%%%%%%%%%%%%%
\section{METHODOLOGY}

\subsection{System Framework}
As shown in Fig.\ref{SystemFramework}, our BVIP guiding system concludes 4 subsystems: Mapping and Localization System, Global Path Planning System, Motion Planning System, and Actuator Controlling System.

Firstly, Mapping and Localization System perceives the environment. The map and the robot's position are obtained by the SLAM algorithm called Cartographer\cite{hess2016real}. Human agent's position is obtained by the camera positioning the AprilTags\cite{wang2016apriltag} on the torso.

Next, Global Path Planning System links the starting point and the final target. We use the Dijkstra algorithm to plan a global path, and smooth it with a series of smoothers introduced by Song et al.\cite{song2019smoothed} We then select waypoints as human's target points using the waypoints selector we proposed. When selecting waypoints, HMP is used to predict human's future speed, and our selected waypoints match the speed trend.

Motion Planning System concludes 2 parts, human motion planner and robot motion planner. 
Human motion planner is used to plan the tension and direction required for human to follow the planned path, while robot motion planner is used to plan the walking speed and direction required to achieve this tension.
Both two planers use model predictive control algorithm.
In human motion planner, we use HMP to predict the optimal robot target position and leash motor target tension.
To enable the robot to reach these target positions after being pulled, RDM predicts the optimal expected velocity for the robot controller in robot motion planner.

Finally, to ensure that the robot reaches the expected velocity, the Quadruped Robot Controller controls the torque of the quadruped robot's joint motors. The Leash Motor Controller is responsible for ensuring that the leash reaches the target tension by controlling the motor input current.
%%%%%%%%%%%%%%%%%%%%%%%%%%%%%%%%%%%%%%%%%%%%%%%%%%%%%%%%%%%%%%%%%%%%%%%%%%%%%%%%
\subsection{Human Motion Predictor}
Precisely predicting different human agents' motions is essential. Du Toit and Burdick pointed out that the lack of explicit human motion prediction results in uncertainty explosion\cite{du2011robot}. This phenomenon may yield the freezing robot problem described by Trautman et al.\cite{trautman2015robot}

Human agent's motion is influenced by the guidance from the guiding robot. We propose to use deep neural networks to regress the sequential relationship between human's velocity and the received tension in the past. In the following paragraph, we will introduce the data collection method, the backbone architectures of our deep neural networks, and the training method of our model.
%%%%%%%%%%%%%%%%%%%%%%%%%%%%%%%%%%%%%%%%%%%%%%%%%%%%%%%%%%%%%%%%%%%%%%%%%%%%%%%%
\begin{table}[t]
\caption{NOMENCLATURE}
\begin{center}
\begin{tabular}{ccc}
\hline
\textbf{Symbol}     & \textbf{Description}                 & \textbf{Dimension}            \\ \hline
$\theta$            & yaw angle                            & scalar                        \\
$\mathbf{x}$        & position                             & $\mathbb{R}^{3\times1}$       \\
$\mathbf{v}$        & velocity                             & $\mathbb{R}^{3\times1}$       \\
$\mathbf{F}$        & tension                              & $\mathbb{R}^{3\times1}$       \\
$W$                 & sequence length of input             & scalar                        \\
$N$                 & prediction horizon of MPC controller & scalar                        \\
$T$                 & the period of the controller         & scalar                        \\
$\Box^h$            & human's state parameter              & -                             \\
$\Box^r$            & robot's state parameter              & -                             \\
$\Box^*$            & expectation                          & -                             \\
$\widetilde{\Box}$  & prediction                           & -                             \\
$\Box_t$            & state parameter at time $t$          & -                             \\ \hline
\end{tabular}
\end{center}
\end{table}
%%%%%%%%%%%%%%%%%%%%%%%%%%%%%%%%%%%%%%%%%%%%%%%%%%%%%%%%%%%%%%%%%%%%%%%%%%%%%%%%
\subsubsection{Data Collection}
Before collecting the data, we mapped the experimental environment and pre-set several paths. We invited 10 volunteers to hold the leash which was attached to the guide robot, who guided them from the starting point to the target. All volunteers were asked to wear blindfolds, earmuffs and our positioning vest. When collecting data, we used a four wheeled robot as its movement is smoother than quadruped robots, resulting in less oscillation and higher data quality.

Our sampling strategy was as follows: (a). The experiment assistant remotely controlled the robot to follow the planned path from the starting point to the target, and adjusted the speed to maintain a certain distance from different human agents. (b). The tension of the leash motor was randomly changed every 10 seconds, ranging from 2N to 20N. (c). The sampling frequency was set to be 50 Hz. (d). The collected data included the position of the robot $\mathbf{x}^r$, the position of the detected AprilTags, the magnitude of the tension $F$, and the direction of the leash $\mathbf{e}_l$.

After collecting the data, we further processed them. We obtained the relative distance of the human agent and the robot $l$ by averaging the positions of the detected AprilTags. Human agent's positions were calculated by adding the relative positions to the robot's positions:
%------------------------------------------------------------------------------
\begin{align}
\mathbf{x}^h=\mathbf{x}^r+l\mathbf{e}_l
\end{align}
%------------------------------------------------------------------------------
The sampled tension magnitudes was decomposed into orthogonal components:
%------------------------------------------------------------------------------
\begin{align}
\mathbf{F}^h=-F\mathbf{e}_l={[F^h_x, F^h_y, F^h_z]}^\intercal
\end{align}
%------------------------------------------------------------------------------
By taking the derivatives of human agent's positions, we obtained human agent's velocities $\mathbf{v}^h$. (d). The data was smoothed and filtered to acquire better quality. (e). The data was segmented into the form:
%------------------------------------------------------------------------------
\begin{align}
(\mathbf{v}_{k-W^h+1:k}^h, \mathbf{F}_{k-W^h+1:k}^h|\mathbf{v}_{k+1}^h)
\end{align}
%------------------------------------------------------------------------------
, which means human agent's velocities and the received tensions in past $W^h$ timesteps were set to be the input of HMP, and the velocity of next timestep were set to be the label.
%%%%%%%%%%%%%%%%%%%%%%%%%%%%%%%%%%%%%%%%%%%%%%%%%%%%%%%%%%%%%%%%%%%%%%%%%%%%%%%%
\begin{figure}[t]
    \centering
    \includegraphics[width=0.2895\textwidth,height=0.24\textwidth]{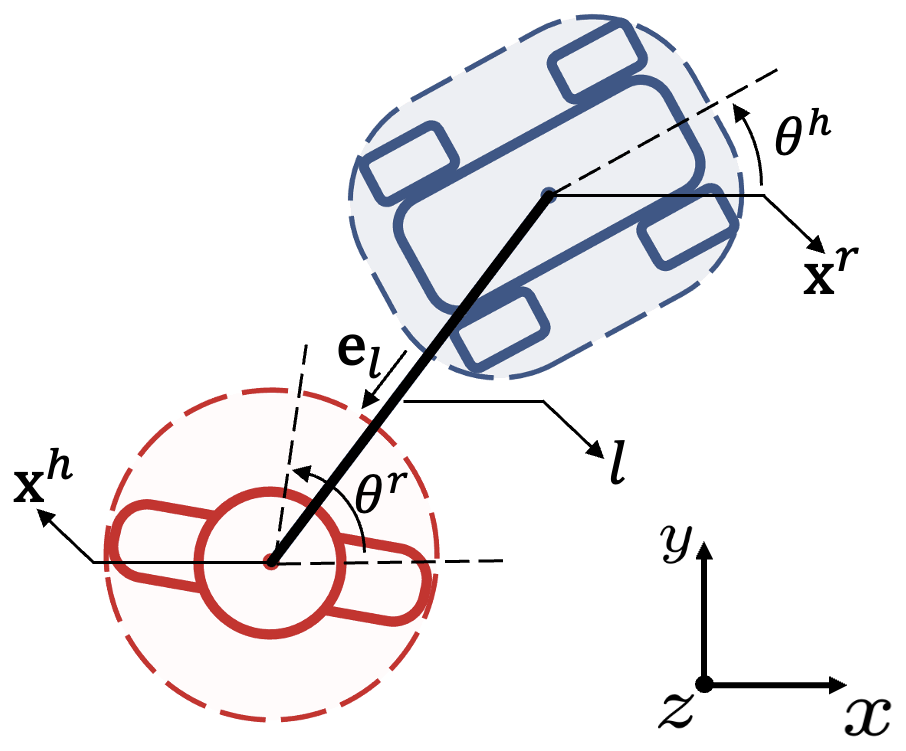}
    \caption{Configuration of the guiding system. The human agent is guided by the robot through a leash.}
    \label{NomenclatureFigure}
\end{figure}
%%%%%%%%%%%%%%%%%%%%%%%%%%%%%%%%%%%%%%%%%%%%%%%%%%%%%%%%%%%%%%%%%%%%%%%%%%%%%%%%
\begin{figure*}[htp]
    \includegraphics[width=0.84\textwidth]{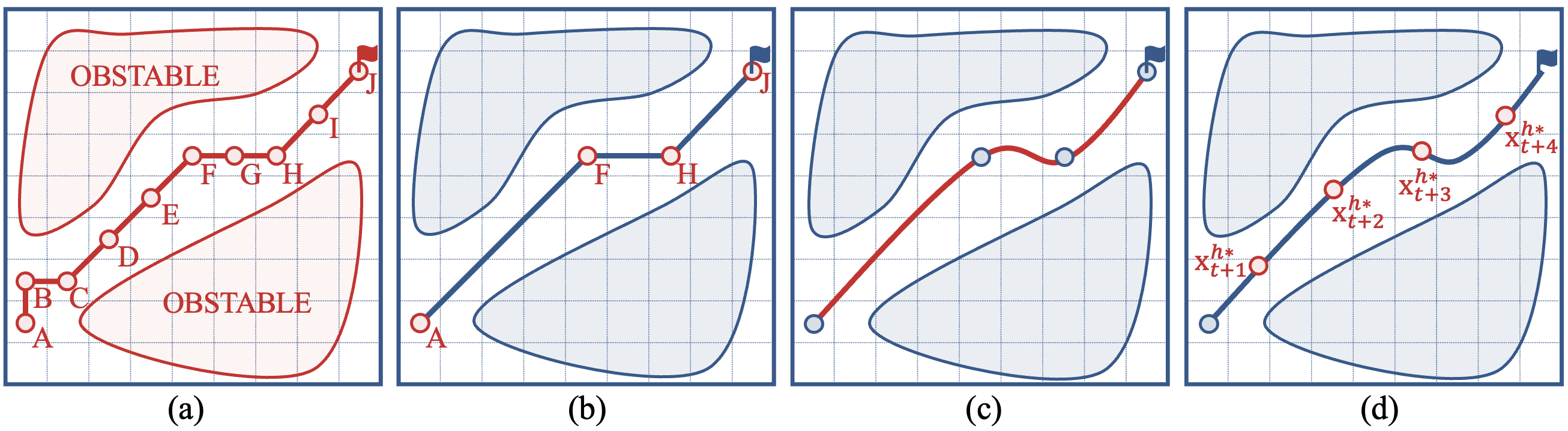}
    \centering
    \caption{Overview of our Global Path Planning System. (a) shows the discrete path based on grid map planned by Dijkstra. (b) is a schematic diagram of LOPS and WRPS removing redundant points. (c) shows the interpolation. (d) is a schematic diagram of our Waypoints Selector selecting waypoints on the analytical path.}
\end{figure*}
%%%%%%%%%%%%%%%%%%%%%%%%%%%%%%%%%%%%%%%%%%%%%%%%%%%%%%%%%%%%%%%%%%%%%%%%%%%%%%%
\subsubsection{Backbone Architecture}
We use 3 well-known architecture varients for the sequence predicting task: CNN, LSTM and TCN.

\noindent\textbf{HMP CNN}: We used the standard convolutional network\cite{lecun1998gradient} to extract features.It contains 2 convolutional layers. The output channel of the convolutional layers are respectively 8 and 16, and the kernel size of each convolutional layer is 3. A fully connected layer is added at the end to regress the extracted features into a velocity vector.

\noindent\textbf{HMP LSTM}: We use a stacked unidirectional LSTM architecture\cite{hochreiter1997long}. It contains 3 layers with 16 units each. A fully connected layer is also added at the end.

\noindent\textbf{HMP TCN}: Our HMP-TCN shares the same architecture with RoNIN-TCN\cite{herath2020ronin} while the dimension of the input and the ouput is revised.
%%%%%%%%%%%%%%%%%%%%%%%%%%%%%%%%%%%%%%%%%%%%%%%%%%%%%%%%%%%%%%%%%%%%%%%%%%%%%%%
\subsubsection{Training Method}
In order to regress the predicted velocity to the real velocity we collected, we designed a loss function:
%------------------------------------------------------------------------------
\begin{align}
    \widetilde{\mathbf{v}}^h_k=f_{\textbf{HMP}}(\mathbf{v}_{k-W^h:k-1}^h, \mathbf{F}_{k-W^h:k-1}^h)
    \label{HMP}
\end{align}
%------------------------------------------------------------------------------
\begin{align}
    \widetilde{\mathbf{x}}^h_{k+1}=\widetilde{\mathbf{x}}^h_k+\widetilde{\mathbf{v}}^h_kT
    \label{HumanPosition}
\end{align}
%------------------------------------------------------------------------------
\begin{align}
loss=L_2(\widetilde{\mathbf{x}}^h, \mathbf{x}^h)
\end{align}
%------------------------------------------------------------------------------

Moreover, we use the optimizer called Adam\cite{kingma2014adam} to optimize the model parameters.
%%%%%%%%%%%%%%%%%%%%%%%%%%%%%%%%%%%%%%%%%%%%%%%%%%%%%%%%%%%%%%%%%%%%%%%%%%%%%%%%
\subsection{Robot Dynamics Model}
Model-based control can generally result in superior performance. However, it is difficult to precisely describe the relationship between the quadruped robot's velocity and the tension it received. On the one hand, Quadruped robot has 12 DoF, which makes the dynamics model become complicated to build manually. On the other hand, quadruped robot has complicated motion controllers, which makes the impact of the tensions which were received in past periods are integrated, to influence its velocity. To accurately establish the model, we use deep neural networks to regress the relationship rather than designing the model manually.
%%%%%%%%%%%%%%%%%%%%%%%%%%%%%%%%%%%%%%%%%%%%%%%%%%%%%%%%%%%%%%%%%%%%%%%%%%%%%%%%
\subsubsection{Data Collection}
Our sampling strategy was as follows: (a). Set the quadruped robot to walk at random constant speeds. (b). Applying forces of varying magnitudes, durations and directions to the quadruped robot using our force controlling device. (c). The sampling frequency was set to be 50 Hz. (d). The collected data includes the velocity of the robot $\mathbf{v}^r$, the control input of the Quadruped Robot Controller $\mathbf{u}^r$, the magnitude of the tension $F$, and the direction of the leash $\mathbf{e}_l$.

After the collection, we decomposed the tension into orthogonal components:
%------------------------------------------------------------------------------
\begin{align}
    \mathbf{F}^r=F\mathbf{e}_l={[F^r_x, F^r_y, F^r_z]}^\intercal
\end{align}
%------------------------------------------------------------------------------

We also filtered the data, and segment the data into the form:
%------------------------------------------------------------------------------
\begin{align}
(\mathbf{v}_{k-W^h+1:k}^r, \mathbf{u}_{k-W^h+1:k}^r, \mathbf{F}_{k-W^h+1:k}^r|\mathbf{v}_{k+1}^r)
\end{align}
%------------------------------------------------------------------------------

Our Robot Dynamics Model shares the 3 well-known architecture variants with our Human Motion Predictor, while the dimension of the input is revised.
%%%%%%%%%%%%%%%%%%%%%%%%%%%%%%%%%%%%%%%%%%%%%%%%%%%%%%%%%%%%%%%%%%%%%%%%%%%%%%%%
\subsection{Waypoints Selector}
The function of Waypoints Selector is to select certain waypoints from the path generated by Dijkstra as human's target positions in next timesteps. The selected waypoints should be correlated with the speed trend of different human agents: the faster the speed of the human agent, the greater the distance between the waypoints should be.

Before selecting the waypoints, we need to transform the discrete path based on the grid map into a smooth analytical path. We used two smoothers called LOPS and WRPS which were proposed by Song et al.\cite{song2019smoothed} to remove redundant points. Then, we used the spline Interpolation Path Smoother (IPS) to transform the remained path points into an analytical path. 

To predict human agent's speed in next timesteps, our Neural-Based Human State Predictor is used:
%------------------------------------------------------------------------------
\begin{align}
    \Vert\widetilde{\mathbf{v}}^h_k\Vert={\Vert}f_{\textbf{HMP}}(\mathbf{v}_{k-W^h:k-1}^h, \mathbf{F}_{k-W^h:k-1}^h)\Vert
\end{align}
%------------------------------------------------------------------------------
We assumed that the tension remains unchanged in the future when predicting human's speed.

Finally, we use our waypoints selector to select waypoints on the smoothed analytical path as human agent's expected positions in next periods. The distances between the selected waypoints satisfy:
%------------------------------------------------------------------------------
\begin{align}
    \Vert\mathbf{x}^{h*}_{k+1}-\mathbf{x}^{h*}_k\Vert=\Vert\widetilde{\mathbf{v}}^h_k{\Vert}T
\end{align}
%%%%%%%%%%%%%%%%%%%%%%%%%%%%%%%%%%%%%%%%%%%%%%%%%%%%%%%%%%%%%%%%%%%%%%%%%%%%%%%
\subsection{Motion Planning System}
\subsubsection{Human Motion Planner}
The function of Human Motion Planner is to use optimization algorithms to solve for the best robot position and leash tension, in order to maximize the probability of human agents reaching the expected position planned by the Waypoints Selector. MPC algorithm consists of three parts: a predictive model, feedback correction, and optimization in range of a temporal window.

The model is already explained in section III.(B). We use deep neural network to predict the relationship of human agent's motion and the received tension according to formula \ref{HMP}. While predicting human agent's velocity in next $M^h$ timesteps, the prediction of the tension $\widetilde{\mathbf{F}}$ was set to be the input when $k>t$ rather than the feedback $\mathbf{F}$.

Then, human agent's position can be predicted according to formula \ref{HumanPosition}, and robot's target position can be predicted according to:
%------------------------------------------------------------------------------
\begin{align}
    \widetilde{\mathbf{x}}^{r*}=\mathbf{\widetilde{x}}^h+l\mathbf{e}_\theta
\end{align}
%------------------------------------------------------------------------------

Next, to acquire the optimal control variables, we minimized the cost function:
%------------------------------------------------------------------------------
\begin{subequations}
\begin{align}
J^h(\widetilde{\mathbf{F}}^h, \widetilde{l}, \widetilde{\theta})=\sum_{k=0}^{M^h-1}&(\Vert\widetilde{\mathbf{x}}^h_k-\mathbf{x}^{h*}_{k}\Vert_{\omega_1} \\
+\Vert\widetilde{\mathbf{F}}_{k+1}-\mathbf{\widetilde{F}}_k\Vert_{\omega_2}+\omega_3&(\Vert\mathbf{\widetilde{F}}_{k+1}\Vert-\Vert\mathbf{\widetilde{F}}_k\Vert)^2 \\
+\omega_4(1-\cos(\widetilde{\theta}_{k+1}-\widetilde{\theta}_k&))+\omega_5(\widetilde{l}_{k+1}-\widetilde{l}_k)^2)
\end{align}
%------------------------------------------------------------------------------
\begin{equation}
    F_{min}<\Vert\widetilde{\mathbf{F}}_k\Vert<F_{max}
\end{equation}
%------------------------------------------------------------------------------
\begin{equation}
    l_{min}<\widetilde{l}_k<l_{max}
\end{equation}
%------------------------------------------------------------------------------
\begin{equation}
    \left\langle\widetilde{\mathbf{F}}_k, \widetilde{\mathbf{F}}_{k+1}\right\rangle\leq\varphi_{\mathbf{F}}
\end{equation}
%------------------------------------------------------------------------------
\begin{equation}
    \left\langle\mathbf{e}_{\widetilde{\mathbf{F}}_k}, \mathbf{e}_{\widetilde{\theta}_k}\right\rangle\leq\varphi_{\theta}
\end{equation}
%------------------------------------------------------------------------------
\begin{equation}
    \Vert\widetilde{\mathbf{x}}_k^h-\mathbf{x}_j^{obs}\Vert{\geq}r^{obs}
\end{equation}
%------------------------------------------------------------------------------
\begin{equation}
    \Vert\widetilde{\mathbf{x}}^{r*}_k-\mathbf{x}^{obs}_j\Vert{\geq}r^{obs}
\end{equation}
\end{subequations}
%------------------------------------------------------------------------------
where $\Vert{\mathbf{x}}\Vert_{\omega}=\frac{1}{2}\omega\mathbf{x}^\intercal\mathbf{x}$. $\omega_1, \omega_2\in\mathbb{R}^{3\times3}, \omega_3, \omega_4, \omega_5\in\mathbb{R}$ are weight coefficients, $F_{min}, F_{max}, l_{min}, l_{max}, \varphi_{\mathbf{F}}, \varphi_{\theta}, r^{obs}$ are the the thresholds.

As for the cost function, minimizing $\sum_{k=0}^{M^h-1}{\Vert\mathbf{\widetilde{x}}^h_{k}-\mathbf{x}^{h*}_k\Vert}$ promises the predicted trajectory $\mathbf{\widetilde{x}}^h$ to approach the planned waypoints $\mathbf{x}^{h*}$; minimizing $\sum_{k=0}^{M^h-1}{\Vert\mathbf{\widetilde{F}}_{k+1}-\mathbf{\widetilde
{F}}_k\Vert}$ and $\sum_{k=0}^{M^h-1}(\Vert\mathbf{\widetilde{F}}_{k+1}\Vert-\Vert\mathbf{\widetilde{F}}_k\Vert)$ promises the direction and magnitude of the force to change smoothly; and minimizing $\sum_{k=0}^{M^h-1}{(1-\cos(\widetilde{\theta}_{k+1}-\widetilde{\theta}_k))}$ and $\sum_{k=0}^{M^h-1}{(\widetilde{l}_{k+1}-\widetilde{l}_k)^2)}$ promises the direction and magnitude of the leash to change smoothly.

Finally, through the optimization in range of a temporal window, we obtained the optimal rope motor force $\mathbf{F}^*$, optimal relative distance $l^*$, and optimal relative yaw angle $\theta^*$. Robot's target positions can be calculated according to:
%------------------------------------------------------------------------------
\begin{align}
    \mathbf{x}^{r*}=\mathbf{x}^{h*}+l^*\mathbf{e}_{\theta^*}
\end{align}
%------------------------------------------------------------------------------
%%%%%%%%%%%%%%%%%%%%%%%%%%%%%%%%%%%%%%%%%%%%%%%%%%%%%%%%%%%%%%%%%%%%%%%%%%%%%%%%
\subsubsection{Robot Motion Planner}
The function of Robot Motion Planner is to plan the best control input for the Quadruped Robot Controller, so that the robot can reach the expected position planned by the Human Motion Planner.

The model is already explained in section III.(C). The model described robot's velocity after being towed:
%------------------------------------------------------------------------------
\begin{align}
    \widetilde{\mathbf{v}}^r_k=f_{\textbf{RDM}}(\mathbf{v}_{k-W^h:k-1}^r, \mathbf{u}_{k-W^h:k-1}^r, \mathbf{F}_{k-W^h:k-1}^r)
\end{align}
%------------------------------------------------------------------------------
While predicting robot's velocity in next $M^r$ timesteps, the prediction of the control input $\widetilde{\mathbf{u}}^r$ was set to be the input when $k>t$ rather than the historic value $\mathbf{u}^r$. Robot's position can be predicted according to:
\begin{align}
    \widetilde{\mathbf{x}}^r_{k+1}=\widetilde{\mathbf{x}}^r_k+\mathbf{v}^r_kT
\end{align}
%------------------------------------------------------------------------------

Next, to acquire the optimal control variables, we minimized the cost function:
%------------------------------------------------------------------------------
\begin{subequations}
\begin{equation}
J^r(\widetilde{\mathbf{u}}^r)=\sum_{k=0}^{M^r-1}{(\Vert\widetilde{\mathbf{x}}^r_k-\mathbf{x}^{r*}_k\Vert_{\omega_6}+\Vert\widetilde{\mathbf{u}}^r_k\Vert_{\omega_7})}
\end{equation}
%------------------------------------------------------------------------------
\begin{equation}
\Vert\widetilde{\mathbf{u}}^r_x\Vert_{\infty}{\leq}u^r_{xmax}
\end{equation}
%------------------------------------------------------------------------------
\begin{equation}
\Vert\widetilde{\mathbf{u}}^r_y\Vert_{\infty}{\leq}u^r_{ymax}
\end{equation}
%------------------------------------------------------------------------------
\begin{equation}
\Vert\widetilde{\mathbf{u}}^r_{\omega}\Vert_{\infty}{\leq}u^r_{{\omega}max}
\end{equation}
\end{subequations}
%------------------------------------------------------------------------------
where $\omega_6, \omega_7\in\mathbb{R}^{3\times3}$ are weight coefficients, $u^r_{xmax}, u^r_{ymax}, u^r_{{\omega}max}$ are the the thresholds.

As for the cost function, minimizing $\sum_{k=0}^{M^r-1}{\Vert\widetilde{\mathbf{x}}^r_k-\mathbf{x}^{r*}_k\Vert}$ promises robot's predicted positions $\widetilde{\mathbf{x}}^r$ to approach the planned target positions $\mathbf{x}^{r*}$; and minimizing $\sum_{k=0}^{M^r-1}{\Vert\widetilde{\mathbf{u}}^r_k\Vert}$ promises the robot to walk efficiently.

Finally, through the optimization in range of a temporal window, we obtained the optimal input $\mathbf{u}^{r*}$ for the Quadruped Robot Controller.
%%%%%%%%%%%%%%%%%%%%%%%%%%%%%%%%%%%%%%%%%%%%%%%%%%%%%%%%%%%%%%%%%%%%%%%%%%%%%%%%
\section{EXPERIMENTS}
\subsection{Experimental Settings}
\begin{figure}[h]
    \centering
    \includegraphics[width=0.48\textwidth,height=0.2288\textwidth]{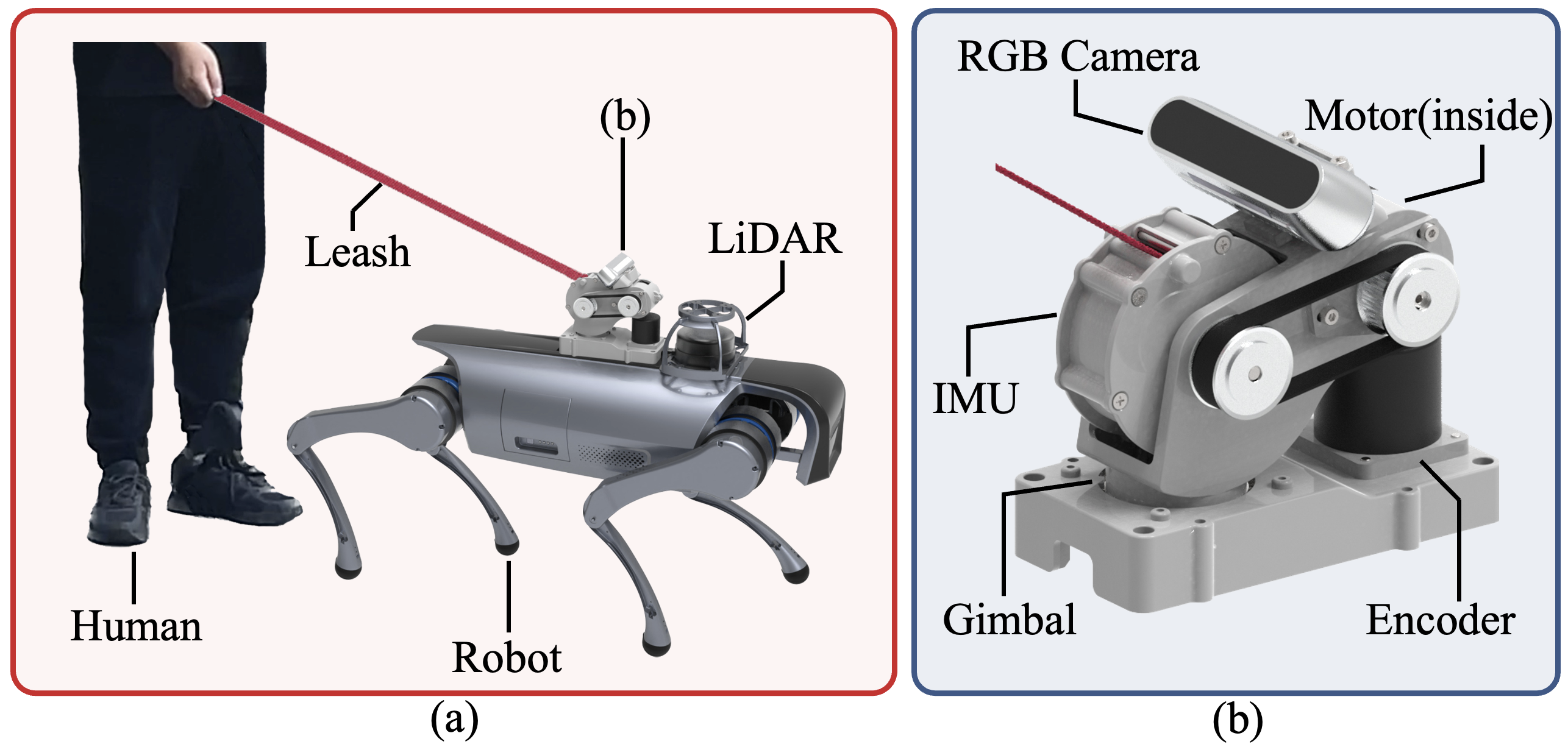}
    \caption{Our hardware platform consists of a quadruped robot, different sensors for perceiving the environment, and actuators to execute the guiding task.}
    \label{HardWare}
\end{figure}

Our hardware platform is shown in Fig.\ref{HardWare}. Our sensors include a LiDAR for SLAM, a RGB camera and encoder for human localization, a force sensor for measuring leash tension, and an IMU for measuring the euler angle of the leash. The actuators include a quadruped robot and a leash motor for traction. The leash motor guides the human agent through the leash.

Robot Operating System(ROS)\cite{quigley2009ros} is used for communication.
%%%%%%%%%%%%%%%%%%%%%%%%%%%%%%%%%%%%%%%%%%%%%%%%%%%%%%%%%%%%%%%%%%%%%%%%%%%%%%%%
%%%%%%%%%%%%%%%%%%%%%%%%%%%%%%%%%%%%%%%%%%%%%%%%%%%%%%%%%%%%%%%%%%%%%%%%%%%%%%%%
\subsection{Results}
\subsubsection{Human Motion Prediction}
To test the performance of our Human Motion Predictor, we compared our HMP model with the linear model proposed by Chen et al.\cite{chen2022quadruped} and the GeoC model proposed by Nanavati et al.\cite{nanavati2019follow}.

As shown in TABLE \ref{HumanTable}, we compare the performance of different models on our dataset using K-fold Cross-Validation. 
We calculate the error and standard deviation between the predicted human velocity and the ground truth velocity for each model. 
The result shows that our models has smaller error and standard deviation, indicating that our models is closer to the ground truth velocity and have a more stable performance. 
Thanks to its more advanced architecture, our HMP-TCN model performed the best among the three variants. 
It can be concluded that our model achieved a significant performance improvement.
%%%%%%%%%%%%%%%%%%%%%%%%%%%%%%%%%%%%%%%%%%%%%%%%%%%%%%%%%%%%%%%%%%%%%%%%%%%%%%%%
\begin{table}[h]
\centering
\caption{The Comparison of Human Motion Predicting Models}
\label{HumanTable}
\begin{tabular}{ccccc}
\hline
\multicolumn{2}{|c|}{\multirow{2}{*}{Model Name}}                       & \multicolumn{1}{c|}{\multirow{2}{*}{Joint Error, $e$}} & \multicolumn{2}{c|}{Velocity Error}                                                       \\ \cline{4-5} 
\multicolumn{2}{|c|}{}                                                  & \multicolumn{1}{c|}{}                                  & \multicolumn{1}{c|}{$e_x$}                  & \multicolumn{1}{c|}{$e_y$}                  \\ \hline
\multicolumn{2}{|c|}{Linear}                                            & \multicolumn{1}{c|}{6.153 (0.683)}                     & \multicolumn{1}{c|}{7.389 (0.592)}          & \multicolumn{1}{c|}{4.917 (0.693)}          \\ \hline
\multicolumn{2}{|c|}{GeoC}                                              & \multicolumn{1}{c|}{0.354 (0.074)}                     & \multicolumn{1}{c|}{0.378 (0.046)}          & \multicolumn{1}{c|}{0.330 (0.057)}          \\ \hline
\multicolumn{1}{|c|}{\multirow{3}{*}{HMP}} & \multicolumn{1}{c|}{CNN}  & \multicolumn{1}{c|}{0.108 (0.295)}                     & \multicolumn{1}{c|}{0.126 (0.261)}          & \multicolumn{1}{c|}{0.090 (0.328)}          \\ \cline{2-5} 
\multicolumn{1}{|c|}{}                      & \multicolumn{1}{c|}{LSTM} & \multicolumn{1}{c|}{4.669 (0.204)}                     & \multicolumn{1}{c|}{7.315 (0.125)}          & \multicolumn{1}{c|}{2.023 (0.344)}          \\ \cline{2-5} 
\multicolumn{1}{|c|}{}                      & \multicolumn{1}{c|}{TCN}  & \multicolumn{1}{c|}{\textbf{0.036 (0.069)}}            & \multicolumn{1}{c|}{\textbf{0.051 (0.053)}} & \multicolumn{1}{c|}{\textbf{0.022 (0.056)}} \\ \hline
\multicolumn{5}{c}{Values are: Avg (StdDev). \textbf{Bold Font} represents best performance.}                                                                                                                                         
\end{tabular}
\end{table}
%%%%%%%%%%%%%%%%%%%%%%%%%%%%%%%%%%%%%%%%%%%%%%%%%%%%%%%%%%%%%%%%%%%%%%%%%%%%%%%%
\begin{table}[h]
\centering
\caption{The Comparison of Robot Dynamics Models}
\label{RobotTable}
\begin{tabular}{ccccc}
\hline
\multicolumn{2}{|c|}{\multirow{2}{*}{Model Name}}                       & \multicolumn{1}{c|}{\multirow{2}{*}{Joint Error, $e$}} & \multicolumn{2}{c|}{Velocity Error}                                                       \\ \cline{4-5} 
\multicolumn{2}{|c|}{}                                                  & \multicolumn{1}{c|}{}                                  & \multicolumn{1}{c|}{$e_x$}                  & \multicolumn{1}{c|}{$e_y$}                  \\ \hline
\multicolumn{2}{|c|}{VDCM}                                              & \multicolumn{1}{c|}{7.152 (0.692)}                     & \multicolumn{1}{c|}{8.169 (0.398)}          & \multicolumn{1}{c|}{6.135 (0.756)}          \\ \hline
\multicolumn{1}{|c|}{\multirow{3}{*}{RDM}} & \multicolumn{1}{c|}{CNN}  & \multicolumn{1}{c|}{3.546 (0.264)}                     & \multicolumn{1}{c|}{4.886 (0.291)}          & \multicolumn{1}{c|}{2.206 (0.276)}          \\ \cline{2-5} 
\multicolumn{1}{|c|}{}                      & \multicolumn{1}{c|}{LSTM} & \multicolumn{1}{c|}{0.258 (0.127)}                     & \multicolumn{1}{c|}{0.337 (0.143)}          & \multicolumn{1}{c|}{0.178 (0.111)}          \\ \cline{2-5} 
\multicolumn{1}{|c|}{}                      & \multicolumn{1}{c|}{TCN}  & \multicolumn{1}{c|}{\textbf{0.042 (0.059)}}            & \multicolumn{1}{c|}{\textbf{0.069 (0.061)}} & \multicolumn{1}{c|}{\textbf{0.014 (0.049)}} \\ \hline
\multicolumn{5}{c}{Values are: Avg (StdDev). \textbf{Bold Font} represents best performance.}
\end{tabular}
\end{table}
%%%%%%%%%%%%%%%%%%%%%%%%%%%%%%%%%%%%%%%%%%%%%%%%%%%%%%%%%%%%%%%%%%%%%%%%%%%%%%%%
\subsubsection{Robot Dynamics Model}
To test the performance of our Robot Dynamics Model, we compared our model with the Velocity Discount Coefficient Matrix(VDCM) model proposed by Chen et al.\cite{chen2022quadruped}

As shown in TABLE \ref{RobotTable}, we also use K-fold Cross-Validation to compare the performance of different models on the dataset. We calculate the error and standard deviation between the predicted robot velocity and the ground truth velocity for different models. Same as the results of NHMP, our models achieve good performance in both accuracy and stability, indicating that our models is effective in predicting robot's velocity.
%%%%%%%%%%%%%%%%%%%%%%%%%%%%%%%%%%%%%%%%%%%%%%%%%%%%%%%%%%%%%%%%%%%%%%%%%%%%%%%%
\subsection{Robot-Human Synchronization}
To test the ability to adapt to individual differences among human agents, we monitor the speeds of both human and robot, and the relative distance between them during the guiding process.

\begin{figure}[h]
    \centering
    \includegraphics[width=0.48\textwidth,height=0.3518\textwidth]{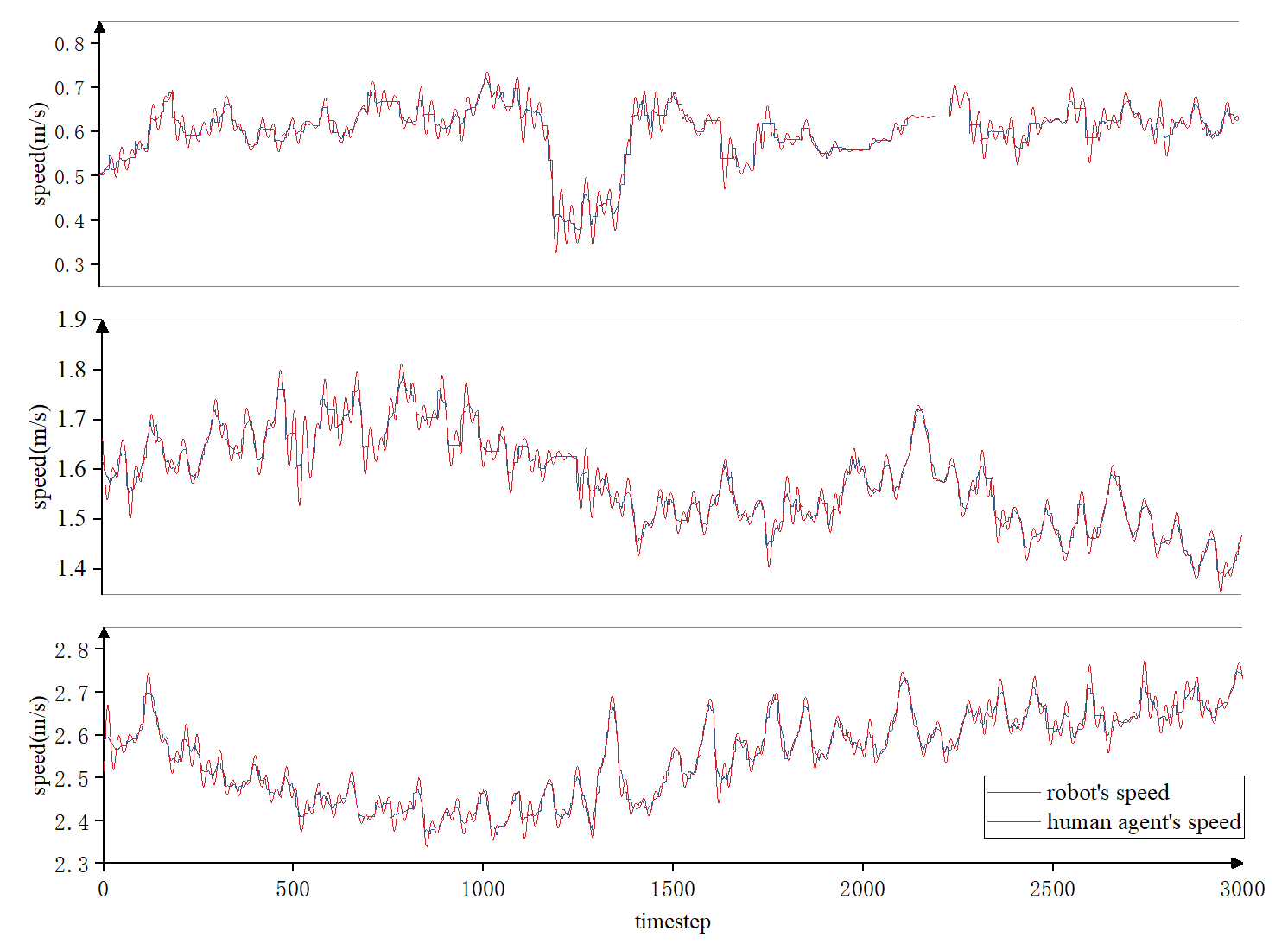}
    \caption{Different human agents are guided by our quadruped robot equipped with our BVIP guiding system. Each of them has their own comfortable walking speed, and our robot is able to keep up with their speed.}
    \label{Synchronization}
\end{figure}

As shown in Fig. \ref{Synchronization}, it can be seen that the robot can maintain a roughly similar speed with different human agents, and the distance between them can be kept stable during the guiding process. It can be concluded that our BVIP guiding system can adapt well to individual differences among different human agents.
%%%%%%%%%%%%%%%%%%%%%%%%%%%%%%%%%%%%%%%%%%%%%%%%%%%%%%%%%%%%%%%%%%%%%%%%%%%%%%%%
%%%%%%%%%%%%%%%%%%%%%%%%%%%%%%%%%%%%%%%%%%%%%%%%%%%%%%%%%%%%%%%%%%%%%%%%%%%%%%%%
\subsection{Application for Real Robot}
We apply the proposed guiding system to a real quadruped robot. 
The quadruped robot platform uses a a robot called WR1 developed by MiLAB Lab, as Fig \ref{realrobot}. 
Test videos will be shown in supplementary materials, and the results show superior guiding performance in different people and at different speeds.

%%%%%%%%%%%%%%%%%%%%%%%%%%%%%%%%%%%%%%%%%%%%%%%%%%%%%%%%%%%%%%%%%%%%%%%%%%%%%%%%
\begin{figure}[h]
    \centering
    \includegraphics[width=0.4\textwidth,height=0.2\textwidth]{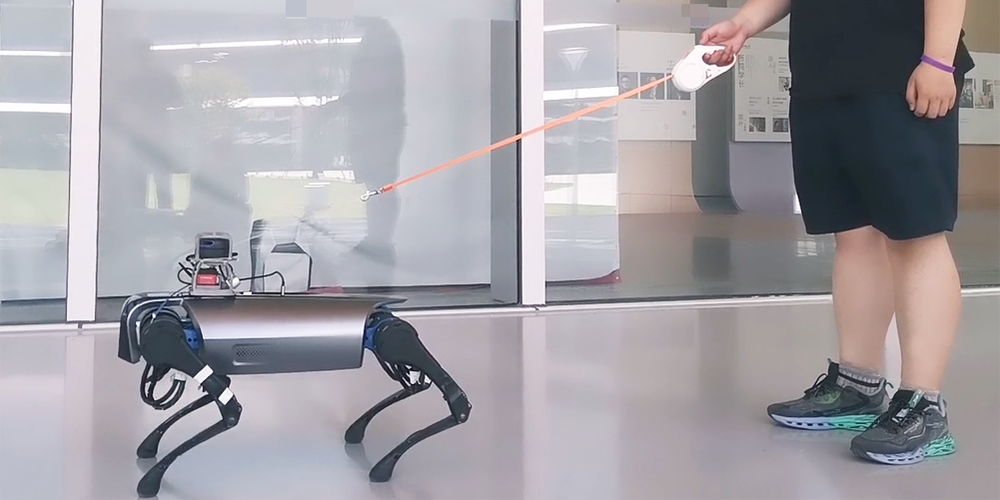}
    \caption{Our BVIP guiding system has been deployed on a robot called WR1 developed by MiLAB Lab.}
    \label{realrobot}
\end{figure}
%%%%%%%%%%%%%%%%%%%%%%%%%%%%%%%%%%%%%%%%%%%%%%%%%%%%%%%%%%%%%%%%%%%%%%%%%%%%%%%%

\section{CONCLUSION}
In this paper, we consider that the differences among human agents require different guiding strategies. 
In our proposed BVIP guiding system, we accurately model the motions of different human agents, and the different effects imposed on the robot, using AI-based HMP and RDM for sequence prediction. 
We compared our models with previous models and demonstrated that our models achieved outstanding performance. 
To further adapt the robot to individual differences, we propose the waypoints selector to correct the global path. 
It was shown that our BVIP robot can synchronize with the speed of different human agents and maintain a stable distance. 
In summary, our system successfully addressed the adaptation problem of individual differences in the BVIP guiding task. 
In future work, we will improve the system so that it can adapt to individual differences without pre-collected data.

%\clearpage
%%%%%%%%%%%%%%%%%%%%%%%%%%%%%%%%%%%%%%%%%%%%%%%%%%%%%%%%%%%%%%%%%%%%%%%%%%%%%%%%
\bibliographystyle{IEEEtran}
\bibliography{references}
%%%%%%%%%%%%%%%%%%%%%%%%%%%%%%%%%%%%%%%%%%%%%%%%%%%%%%%%%%%%%%%%%%%%%%%%%%%%%%%%
\end{document}